\documentclass[a4paper,twoside]{article}

\usepackage{epsfig}
\usepackage{subcaption}
\usepackage{calc}
\usepackage{amssymb}
\usepackage{amstext}
\usepackage{amsmath}
\usepackage{amsthm}
\usepackage{multicol}
\usepackage{pslatex}
\usepackage{apalike}
\usepackage{dsfont}
\usepackage{dblfloatfix} 
\usepackage{SCITEPRESS}     

\begin{document}

\title{The Max-Cut Decision Tree: Improving on the Accuracy and Running Time of Decision Trees}

\author{\authorname{Jonathan Bodine and Dorit S. Hochbaum}
\affiliation{University of California, Berkeley}
\email{jonathan.m.bodine@berkeley.edu, hochbaum@ieor.berkeley.edu}
}


\abstract{Decision trees are a widely used method for classification, both by themselves and as the building blocks of multiple different ensemble learning methods. The Max-Cut decision tree involves novel modifications to a standard, baseline model of classification decision tree construction, precisely CART Gini. One modification involves an alternative splitting metric, maximum cut, based on maximizing the distance between all pairs of observations belonging to separate classes and separate sides of the threshold value. The other modification is to select the decision feature from a linear combination of the input features constructed using Principal Component Analysis (PCA) locally at each node. Our experiments show that this node-based localized PCA with the novel splitting modification can dramatically improve classification, while also significantly decreasing computational time compared to the baseline decision tree. Moreover, our results are most significant when evaluated on data sets with higher dimensions, or more classes; which, for the example data set CIFAR-100, enable a 49\% improvement in accuracy while reducing CPU time by 94\%. These introduced modifications dramatically advance the capabilities of decision trees for difficult classification tasks.}

\onecolumn \maketitle \normalsize \setcounter{footnote}{0} \vfill

\section{\uppercase{Introduction}}
\label{sec:introduction}

\noindent Decision trees are a widely used method for classification, both by themselves and as the building blocks of multiple different ensemble learning methods. A standard approach for the construction of decision trees utilizes the Classification and Regression Trees (CART) method \cite{breiman1984classification}. These decision trees are constructed by selecting a threshold value on some feature to separate the observations into two branches, based on some evaluation method. This process continues recursively until a preset stopping criterion is reached. Throughout our analysis, we use as the stopping rule, either the node having only one class present or no split existing such that both sides of the branch contain at least one observation. 

One of the standard methods for evaluating the different splits is the Gini Impurity, which was provided as one of the methods in the original CART methodology \cite{breiman1984classification}. The definition of Gini Impurity is provided as equation 1, where $p_c$ is the fraction of observations that belong to class $c$.
\begin{equation}\label{eq1}
    \sum_{c} p_c \left(1 - p_c\right)
\end{equation}
A threshold -- which is a value in the range of one of the features' values such that the observations are partitioned into two subsets, the samples whose values for that feature are less than the threshold, and those whose values are greater than the threshold -- is selected such that the average Gini Impurity of the resulting two subsets, weighted by their cardinality, is minimized. We will use this methodology as the baseline against which to compare our novel splitting procedure.  

Our splitting procedure is motivated by a shortcoming of the Gini Impurity, in that it does not take into account the distance of the split, just the partition induced by the split. We view this as a weakness since given two options for a threshold that both partition the observations into the same ratios and same sizes, Gini Impurity would show no preference regardless of the distance between these classes.  Moreover, we believe that this distance information is meaningful and preferable, as it produces thresholds that separate the classes further apart in space, resulting in decreased susceptibility to over-fit.  

Our splitting procedure takes inspiration from the Max Cut, which in general is NP-complete. Our use of Max Cut is in one dimension, based on distances with respect to a single feature, restricted to a single threshold, and therefore polynomial-time solvable. Indeed, we show in the appendix a linear-time implementation given the sorted observations. We select a threshold that seeks to maximize the sum of the weighted arcs between the two subsets of the partition induced by that threshold, where the arcs' weights are determined by the distance in the specified feature of the two nodes if those nodes are from different classes and zero otherwise. Using $x_{i,j}$ to represent the value of feature $j$ for observation $i$, $y_i$ to represent the class of observation $i$, and $\theta$ to represent the threshold value, we are solving the optimization problem specified in equation 2.
\begin{equation}\label{eq2}
    \max_{\theta} \sum_{\{i | x_{i,j} \leq \theta\}} \sum_{\{k | x_{k,j} > \theta\}} \mathds{1}_{y_{i} \neq y_{k}}\left|x_{i,j} - x_{k,j}\right|
\end{equation}
Given the fact that the observations can be sorted based on any feature in $O \left(n \log n\right)$ and the linear-time implementation of Max Cut given this fact we have an $O \left(n \log n\right)$ algorithm for solving this problem on any given feature of the feature vector, where $n$ is the number of observations. It is important to note that this is the same asymptotic time complexity achieved when implementing the Gini Impurity method. 

Both of these methods rely on the fact that there is a finite, polynomial, number of thresholds to consider, in fact it is linear in the number of samples in the set considered, namely for every feature the threshold between each successive unique value of that feature. Therefore, feature construction can have a significant impact on the performance of a decision tree. Moreover, how the information is represented can affect the performance. Specifically, if the feature vector, $\mathbf{x}_{i}$, is modified through a change of bases, the splits of the decision tree can be changed. This motivates the goal of finding a good basis in which to represent the feature vector. 

One common way to find the natural bases to represent the feature vector in, is through the use of Principal Component Analysis (PCA) first proposed by Karl Pearson F.R.S. \cite{doi:10.1080/14786440109462720}. PCA iteratively finds the direction that accounts for the most variance in the data, given that it is orthogonal to all the directions previously found. The use of PCA as a preprocessing step when utilizing decision trees is not new. We, however, propose that it be used throughout the construction of the decision tree at every node instead. This is motivated by the idea that distributions could be different in different subspaces of the feature space. Therefore, what is a meaningful direction overall, might not be that meaningful in specific sub-spaces. To account for this, we introduce the following two methods for finding locally meaningful directions. 

The first method that we consider is {\bf Node Features PCA}, where at every decision node, PCA is performed on the original feature vectors using only the observations that have made it to that node in order to find the local principal components. These local principal components are then used as the input features to find the optimal threshold, instead of using the original features. 

The second method, comes from the idea that it is not the directions that best describe the data, but rather the difference between the classes that are important, and is referred to as {\bf Node Means PCA}. Much like Node Features PCA, this algorithm is used at every decision node. The difference is in what points get feed into the PCA algorithm. Specifically, the first step is to think of the data set in the one-vs-rest context, locally calculating the mean position of each of the possible `rest' collections. As an example if there are locally 3 points $(1,0,0),\ (0,1,0)$, and $(0,0,1)$ each belonging to classes A, B, and C respectively, then the following points would be generated $(0,0.5,0.5)$, $(0.5, 0, 0.5)$, and $(0.5,0.5,0)$. These newly generated points are then passed into the PCA algorithm to find a new basis. The original features are then transformed into this new basis, it is important to note that this will necessarily result in a dimensionality reduction in the case where the number of classes locally present is less than the dimension of the original feature vector. The features in this new basis are then used as the input features to find the optimal threshold, instead of using the original features. 


\setcounter{table}{1}

\begin{table*}[hb!]
\caption{Algorithms' nomenclature.}\label{tab:2} \centering
\begin{tabular}{|c|cc|}
  \hline
  & Gini & Max Cut \\
  \hline
  Original & Gini Features & Max Cut Features \\
  Global PCA & Gini Pre PCA Features & Max Cut Pre PCA Features\\
  Local PCA & Gini Node Features PCA & Max Cut Node Features PCA\\
  Local Means PCA & Gini Node Means PCA & Max Cut Node Means PCA\\
  \hline
\end{tabular}
\end{table*}

To understand how these changes affect a decision tree's performance, we performed an extensive experimental analysis of 8 different decision tree algorithms listed in Table 2, each using different combinations of the methods discussed above (section 2). Initially performed was an analysis on more than 20,000 synthetic data sets (section 3.1) with training set sizes ranging between 100 and 300,000, and a testing set always fixed to 100,000 observations. We first considered binary classification problems (section 3.1.2), and then considered decenary classification problems (section 3.1.3).  After validating that the modification produced statistically significant improvements to the baseline Gini Impurity CART model, these results were then validate on real-world data sets (section 3.2). These modifications provide dramatic improvements both in accuracy and computational time, with these advantages increasing as the dimensionality of the data or the number of classes increases (section 4). The significance of these results is demonstrated here with the example of CIFAR-100 \cite{Krizhevsky09learningmultiple}, which has 100 classes, 3,072 dimensions, and 60,000 total observations of which 48,000 are used for training. In this case, our algorithm Max Cut Node Means PCA (see section 2) resulted in a 49.4\% increase in accuracy performance compared to the baseline CART Gini model while simultaneously reducing the CPU time required by 94\%. 

\section{\uppercase{Algorithms}}

\noindent In our analysis, we consider a total of 8 different algorithms. Each of these algorithms was implemented in the Python programming language \cite{10.5555/1593511}. In order to improve the run time of these algorithms both the Gini Impurity and the Max Cut optimal threshold calculation for a given feature were implemented as NumPy \cite{oliphant2006guide} vector operations without the use of any Python for loops. Moreover, both utilize efficient $O \left(n \log n\right)$ implementations. We remark that both the Gini Impurity and Max Cut metric were implemented so that in the event of a tie, they favored more balanced splits, and the split was chosen to be the mean value between the two closet points that should be separated. Other than these subroutines, no difference existed between the implementations of Gini and Max Cut based trees in order to make time-based comparisons as consistent as possible. The scikit-learn \cite{scikit-learn} package was used to perform the PCA. 

As a result of Gini and Max Cut having the same time complexity, the following results on the time complexity of each algorithm hold regardless of choice in evaluation criterion. The time complexities for computing the optimal split at each node are given in Table 1, where $n$ is the number of observations, $d$ is the number of input features, and $p$ is the number of principal components considered. One should note that these are calculated under the following two assumptions. First, that the number of classes is fixed and smaller than the dimension of the feature vector. Second, that $d \leq n$. Again, under this assumption that $d \leq n$, one can see that the computational complexity of these algorithms in ascending order is Node Means PCA, Baseline, Node Features PCA. 

\setcounter{table}{0}

\begin{table}[h]
\caption{Time Complexity of Node Splitting.}\label{tab:1} \centering
\begin{tabular}{|c|c|}
  \hline
  Algorithm & Time Complexity \\
  \hline
  Baseline & $O(d n \log n)$ \\
  Node Features PCA & $O(n^2 p)$ \\
  Node Means PCA & $O(d^2 + n \log n)$ \\
  \hline
\end{tabular}
\end{table}

\setcounter{table}{2}

When looking at the results of our experimental analysis, one may find that the baseline model took longer than expected, given their prior experiences utilizing the implementations available in open source packages, such as that found in scikit-learn \cite{scikit-learn}. We acknowledge this fact; however, we believe that our implementation provides a cleaner comparison considering they were implemented with similar optimizations; where had we imported a package for the baseline model it would have had an advantage due to the significant efforts put into optimizing the package's code, masking differences caused by the underlying algorithms. The code that we used has not been made publicly available since it does not represent the levels of optimization needed for a production-grade package, merely serving the purpose of comparing these different proposed implementations; however, the code will be made available upon request.  

Each of the algorithms is determined by:
\begin{enumerate}
    \item Split criterion, which is Gini or Max Cut.
    
    \item Which feature vector to choose: Original, Global PCA, Local PCA, or Local Means PCA.
\end{enumerate}
Table 2 provides the nomenclature we use to refer to each of these different combinations.
Each of these algorithms was put through the same rigorous experimentation using both synthetic and real-world data sets to understand the advantages and disadvantages of each. The results of these experiments will now be discussed in the next section.

\section{\uppercase{Results}}

\subsection{Synthetic Data Sets}

\noindent We first consider how each of the eight algorithms performed on synthetic data sets. The use of synthetic data sets provided us with some key advantages compared to only testing the algorithms on real world data sets. First, it allowed us to generate an arbitrarily large number of independent data sets allowing us to draw more statistically meaningful conclusions, with results for more than 20,000 data sets. Second, we were able to generate an arbitrary number of training and test examples allowing us to easily generate data sets ranging in different sizes between 100,100 and 400,000 observations, allowing us to gain greater insights, into the algorithms' performances on large data sets, than we could have achieved using only the real world data sets. Lastly, it allowed us to have fine-grained control of the data sets different characteristics.  

\subsubsection{Experimental Design}

\noindent To generate the synthetic data sets, we used the tool datasets.make\_classification provided by scikit-learn \cite{scikit-learn}. We set it such that we only analyzed the case where each class would have one cluster. These clusters were then centered around the vertices of a hyper-cube of a specified dimension (one of the characteristics that we modify). From here, random points are generated from a normal distribution around each clusters' center. These points are then assigned to that clusters' corresponding class, except in 1\% of cases where the class is randomly assigned. The features were then randomly linearly combined within each cluster to generate covariance. We kept the number of redundant and repeated features to zero throughout our experiments. At this point, meaningless dimensions are added to the feature vector (the number of which is a second characteristic that we control). Finally, the features were randomly shifted and then scaled.

Throughout our experiments, we made observations on how the following four factors changed the Accuracy, Wall Clock Time, and the number of leaves. The factors that we controlled were:
\begin{enumerate}
    \item The number of training examples; it is important to note that since we were able to generate arbitrarily many data points for each data set, we were able to have arbitrarily large testing sets; therefore, we chose to keep the size of the testing set consistent throughout at 100,000.  For example, if the training set were set to 10,000 examples, we would generate a data set of 110,000 data points, then randomly select 10,000 for training and keep the remaining 100,000 for testing. 
    
    \item The number of classes.
    
    \item The number of informative dimensions, referred to as the Dimension of Data in our results.
    
    \item The number of meaningless dimensions, referred to as the Noise Dimension in our results. 
\end{enumerate}

For each of the settings of the above parameters that we tried out, we evaluated the models' results on 30 independently generated data sets. For each of these data sets,  the models were presented with the same training/testing examples after the data set had been standardized (zero mean and unit variance) using the scikit-learn StandardScaler \cite{scikit-learn} fit to the training examples. Standardization was used to make the distances in each feature more comparable, which is important in the PCA analysis as well as the Max Cut. Note that both the Gini Impurity method and the orientation of the original feature vector are invariant under this transformation, meaning the baseline algorithm's performance should also be invariant; therefore, this decision could not have artificially inflated our results when compared to the baseline model.   

Using multiple independent data sets, we can calculate the significance of our results using a one-tailed Paired Sample T-Test. If a specific algorithm had the highest average accuracy, we considered the seven null hypotheses that there was no difference in the performance of the specific algorithm compared to each of the others, and the seven respective alternative hypotheses that the specific algorithm outperformed each of the respective other algorithm. We then used the Holm-Bonferroni Method \cite{10.2307/4615733} to correct for the fact that we were doing multiple hypothesis testing and marked where all the null hypotheses were rejected on the accuracy graph with `//' markings at the 5\% significance level. Likewise, when the mean was less than the highest average accuracy, we instead looked at the seven respective alternative hypotheses that the specific algorithm performed worse than each of the respective other algorithm. Again using the Holm-Bonferroni Method \cite{10.2307/4615733}, we marked on the graph with `\textbackslash\textbackslash' markings when at least one of the null hypotheses was rejected at the 5\% level. 

All of these experiments were run on server nodes that each contained two Intel Xeon E5-2670 v2 CPUs for a total of 20 cores (no hyper-threading was used). Each decision tree was constructed using a single core and thread, meaning that up to 20 experiments were running simultaneously on a single node. 

\subsubsection{Binary Classification}

The first set of experiments was to evaluate how the accuracy and run time of each of these algorithms varied based on the size of the training set as well as the number of informative dimensions. For this experiment, we set the number of noise dimensions to be equal to zero. The results of these experiments are summarized in Figure 1 and Figure 2.

\begin{figure*}[!h]
  \centering
   {\epsfig{file = 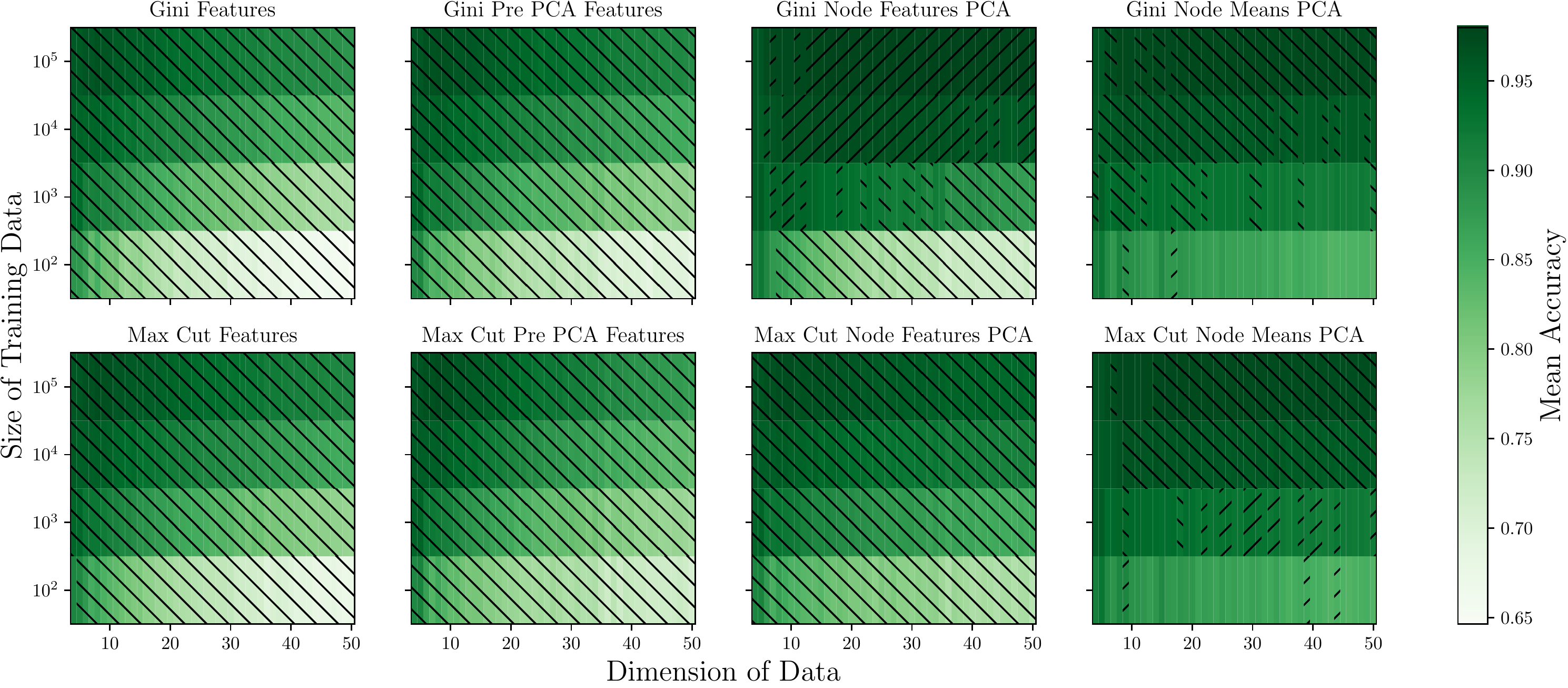, width = 450pt}}
  \caption{Mean accuracy of binary classification on synthetic data, evaluated on test sets of 100,000.}
  \label{fig:Accuracy of Binary Classification on Synthetic Data}
\end{figure*}

\begin{figure*}[!h]
  \centering
   {\epsfig{file = 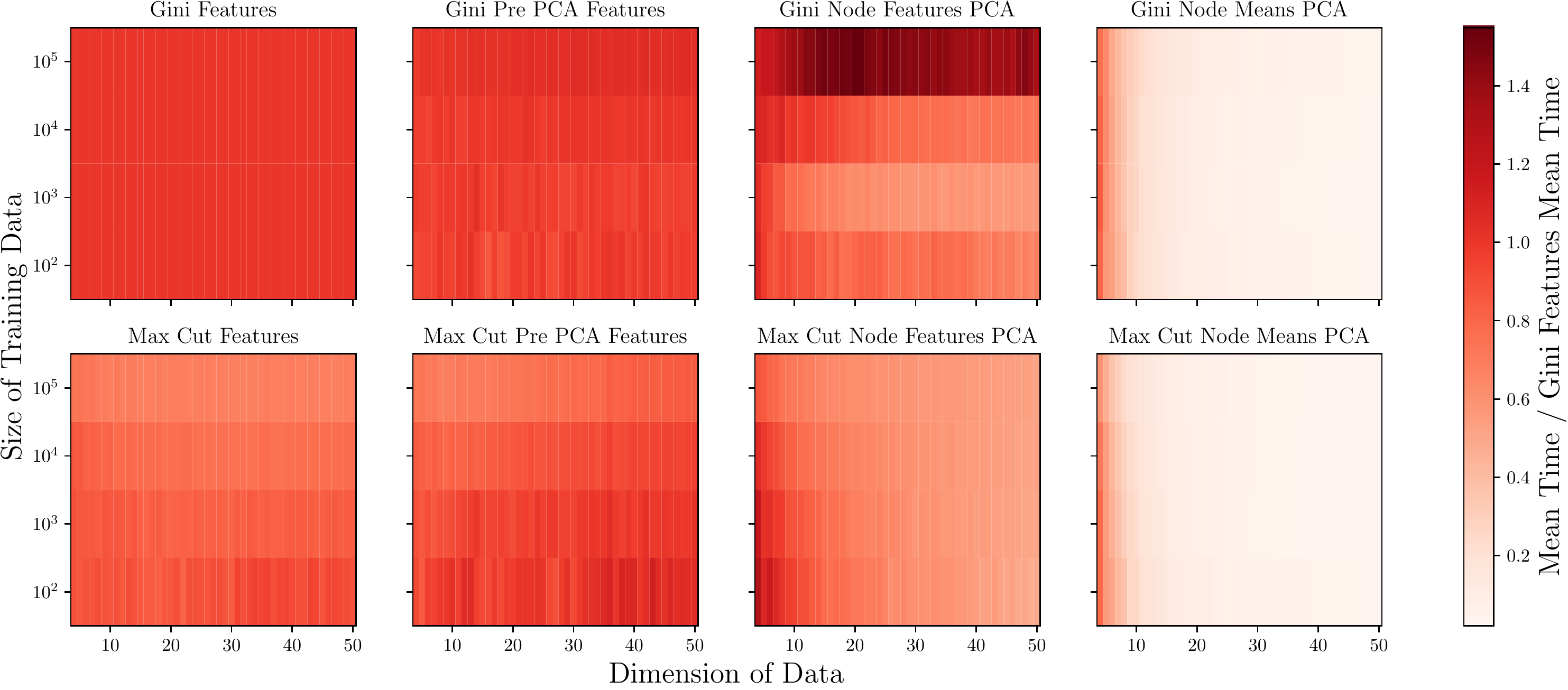, width = 450.pt}}
  \caption{Ratio of mean wall clock time compared to Gini Features baseline for binary classification decision tree construction on synthetic data.}
  \label{fig:Time of Binary Classification on Synthetic Data}
\end{figure*}

Looking at Figure 1, it becomes apparent that the Localized PCA dramatically improves the accuracy when compared to the baseline, as well as when PCA is applied in preprocessing. Moreover, the prevalence of `\textbackslash\textbackslash' markings in the Gini Features, Gini Pre PCA Features, Max Cut Features, Max Cut Pre PCA Features, and Max Cut Node Features PCA indicate with statistical significance that these are not going to be the best algorithm to choose on average. Therefore, we can already conclude that our contributions have dramatically improved upon the baseline model. The consideration then turns to between Gini Node Features PCA, Gini Node Means PCA, and Max Cut Node Mans PCA.  

We first considered Gini Node Means PCA vs. Max Cut Node Means PCA. When looking at Figure 1 and Figure 2, there are no significant trends in their comparison that can be seen, except Max Cut Node Means PCA being the better choice more often, for smaller training sets. We then look at the ratio Max Cut Node Means PCA / Gini Node Means PCA for each of the 5,640 data sets in this experiment. From this analysis, we found the mean ratio to be 1.0003 and the sample standard deviation to be 0.0087, implying a 95\% confidence interval for the mean ratio of $(1.00008, 1.00053)$. This implied that Max Cut Node Means PCA had a slight advantage over the Gini Node Means PCA when considering the average ratio. We, therefor, decided to present the comparison between Gini Node Features PCA vs. Max Cut Node Means PCA. It is important to note that the same patterns emerge if Gini Node Means PCA was used instead, the results are not presented for brevity, as the significant differences come from the choice of finding local directions.   

When considering Gini Node Features PCA vs. Max Cut Node Means PCA, the first pattern that emerges, from Figure 1, is Gini Node Features PCA was the better choice for large training sets (based on accuracy) and Max Cut Node Means PCA was the better choice for smaller training sets. This pattern of Node Means PCA out-performing for harder data sets can be seen in Figure 3, which compares the two algorithms, in the cases where 1,000 training samples were used, where Node Means PCA, after a small dip for low dimensional data, increasingly out-performs the Node Features PCA algorithm as the dimensionality increases. This lower susceptibility to the curse of dimensionality makes sense when one considers the bias-variance trade-off. Node Means PCA uses averaging, thereby reducing the noise. However, it is then only able to look at fewer directions than Node Features PCA due to the dimensionality reduction caused by the number of classes being less than the dimension of the feature space. In fact, in the binary case, the Node Means PCA algorithm is only ever looking at two directions to select a threshold since only two vectors are used when performing the PCA.

\begin{figure}[!h]
  \centering
   {\epsfig{file = 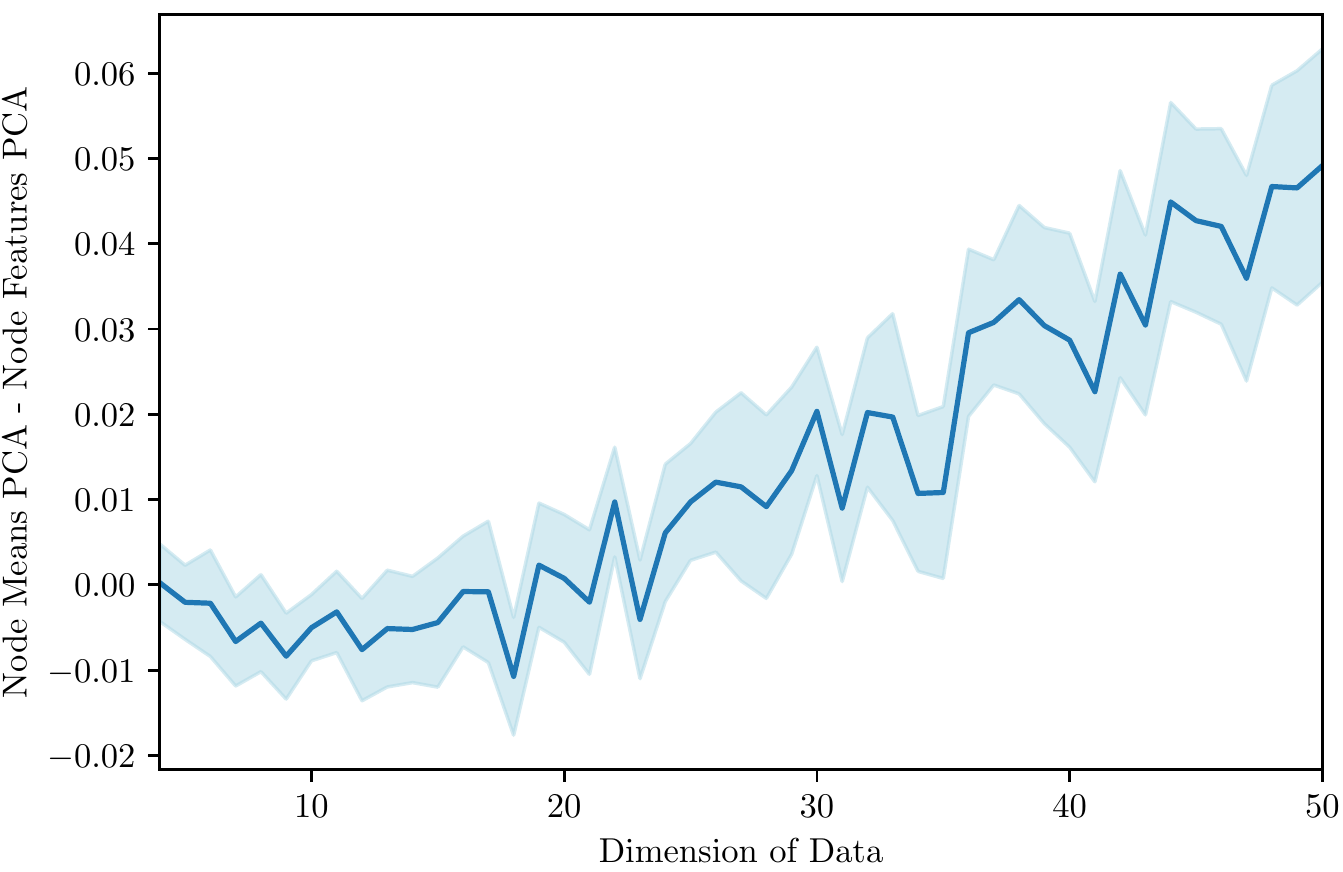, width = 200.pt}}
  \caption{Mean Max Cut Node Means PCA Accuracy minus Mean Gini Node Features PCA Accuracy with 95\% confidence interval using the 1,000 Training Sets.}
  \label{fig:3}
\end{figure}

It is also important to note that the level to which Node Means PCA under-performs is notably less than the level to which it can out-preform. This phenomenon can also be seen clearly in Figure 4 which contains all 5,640 ratios that can be constructed from this experiment and shows the significant skew. Thus, when Node Means PCA out-performs, it does so significantly, but when it under-performs it does so less dramatically. To further demonstrate this point, conditioned on the ratio being greater than 1 (which happens 49.6\% of the time), the mean is 1.075 and the max is 1.430 as compared to when the ratio is conditioned to be less than 1 (which happens 50.4\% of the time), the mean is 0.989 and the minimum is 0.915. This shows that Node Means PCA improves the upside, without creating too much downside, when comparing accuracy. However, there is another consideration to take into account that will demonstrate the dominance of the Node Means PCA method. 

\begin{figure}[!h]
  \centering
   {\epsfig{file = 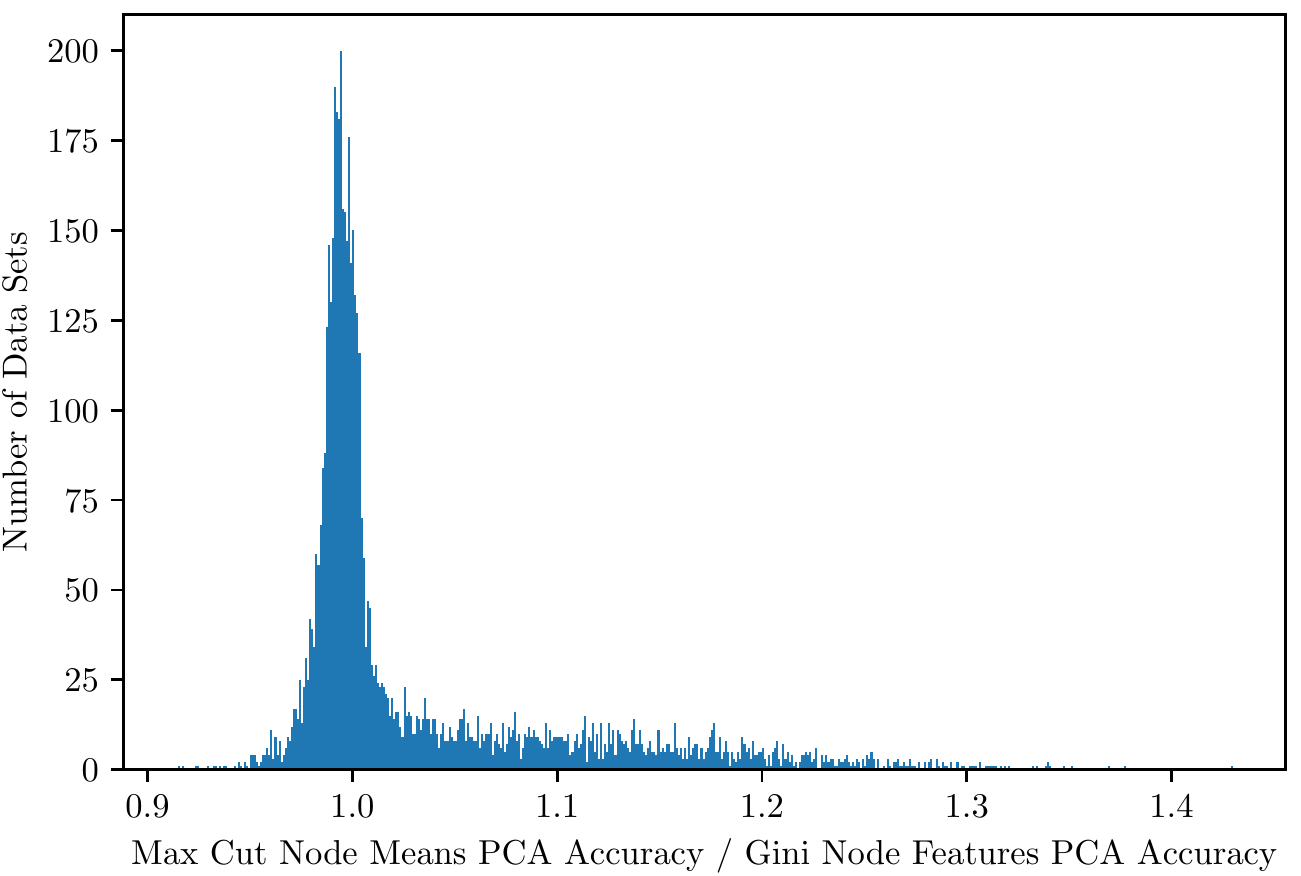, width = 200.pt}}
  \caption{Histogram of Max Cut Node Means PCA Accuracy / Gini Node Features PCA Accuracy.}
  \label{fig:4}
\end{figure}

When examining the results of Node Means PCA vs. Node Features PCA, Figure 2 shows that Node Means PCA has significantly better properties when it comes to computational cost. This fact makes sense, given the analysis of the run times for splitting a node given in section 2. The significance of these results, however, might be more than what would be expected in the Node Means PCA case and less in the Node Features PCA case with the only time where Gini Node Features was noticeably worse is in the 100,000 training set region, which, significantly, is also when it marginally out-performs Node Means PCA. These differences can be explained by the fact that the total run time of constructing a tree is also based on its size. The mean number of leaves across all data sets is provided in Table 3.
\begin{table}[h]
\caption{Mean Number of Leaves in Decision Tree.}\label{tab:3} \centering
\begin{tabular}{|c|c|}
  \hline
  Algorithm & Mean \# of Leaves \\
  \hline
  Baseline & $1129$ \\
  Gini Node Features PCA & $418$ \\
  Gini Node Means PCA & $429$ \\
  Max Cut Node Means PCA & $448$ \\
  \hline
\end{tabular}
\end{table}

Table 3 shows that the average number of leaves is significantly less for Gini Node Features PCA, Gini Node Means PCA, and Max Cut Node Means PCA when compared to the Baseline model, this reduction can help account for the extra observed computation efficiency of the algorithms. In Figure 2 it can be seen that this extra bonus eventually gets balanced out in the Gini Node Features Case for large enough training sets. Therefore, taking into account the costs and benefits related to computational time, accuracy, and robustness to higher dimensional problems, the Node Means PCA algorithm performs the best with a slight preference towards the use of the Max Cut metric. 

The second experiment that was performed considered the addition of noise features. The idea behind this was to see if the Localized PCA algorithms would get tripped up by the noise.  For this experiment the training set sizes were fixed to 10,000 and then we ran 30 runs of each combination of meaningful dimensions in $\{5,10,15,...,50\}$ and noise dimensions in $\{0,5,10,...,50\}$. The major result of this experiment was to discover that localized PCA is susceptible to confusion in extreme cases. When the real dimension was five, and the noise dimension was greater than or equal to 15, our modifications were no longer better than the baseline, with the baseline out-performing ours. However, this represents the cases where noise features account for 75\% or more of the available features, and the number of informative features was small. Therefore, we do not see this as a fundamental problem with our algorithm and move onto the next set of experiments. 

\subsubsection{Decenary Classification}

The next experiment was to see how the results changed when considering the case where there are ten instead of two classes. Utilizing the same steps as in section 3.1.2 one can see how different training set sizes, informative dimensions, and noise dimensions affect the algorithms' performance. The main results that vary the training data size and the underlying dimension of the data can be seen in Figure 5 and Figure 6. One important note is that these plots do not use a perfect logarithmic scale for the y-axis as the values for each horizontal bar are $10^2,\ 10^3,\ 10^4,\ 10^5,\ 2 \times 10^5,\ \text{and } 3 \times 10^5$ (note that the last two are not $10^6$ and $10^7$ as one might initially suspect).  

\begin{figure*}[!h]
  \centering
   {\epsfig{file = 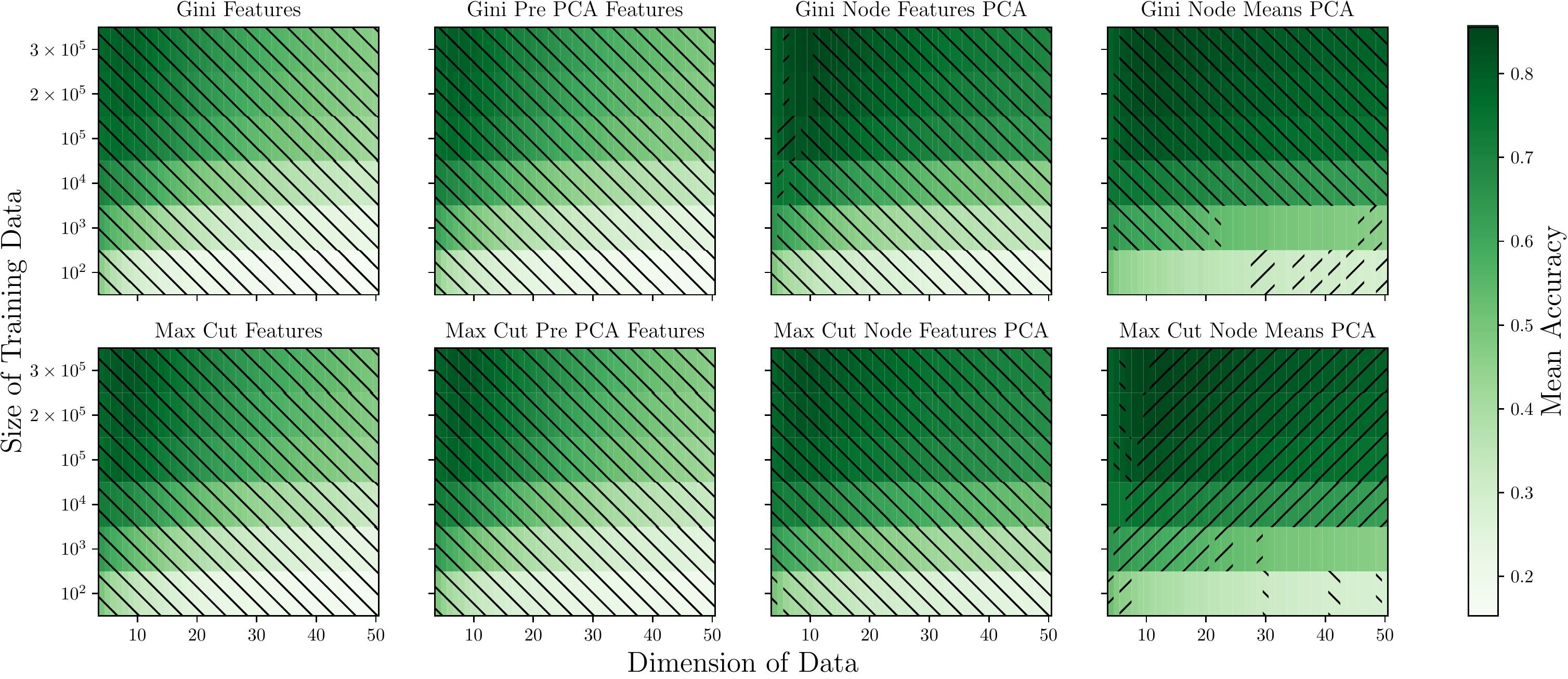, width = 450pt}}
  \caption{Mean accuracy of decenary classification on synthetic data, evaluated on test sets of 100,000.}
  \label{fig:Accuracy of Decenary Classification on Synthetic Data}
\end{figure*}

\begin{figure*}[!h]
  \centering
   {\epsfig{file = 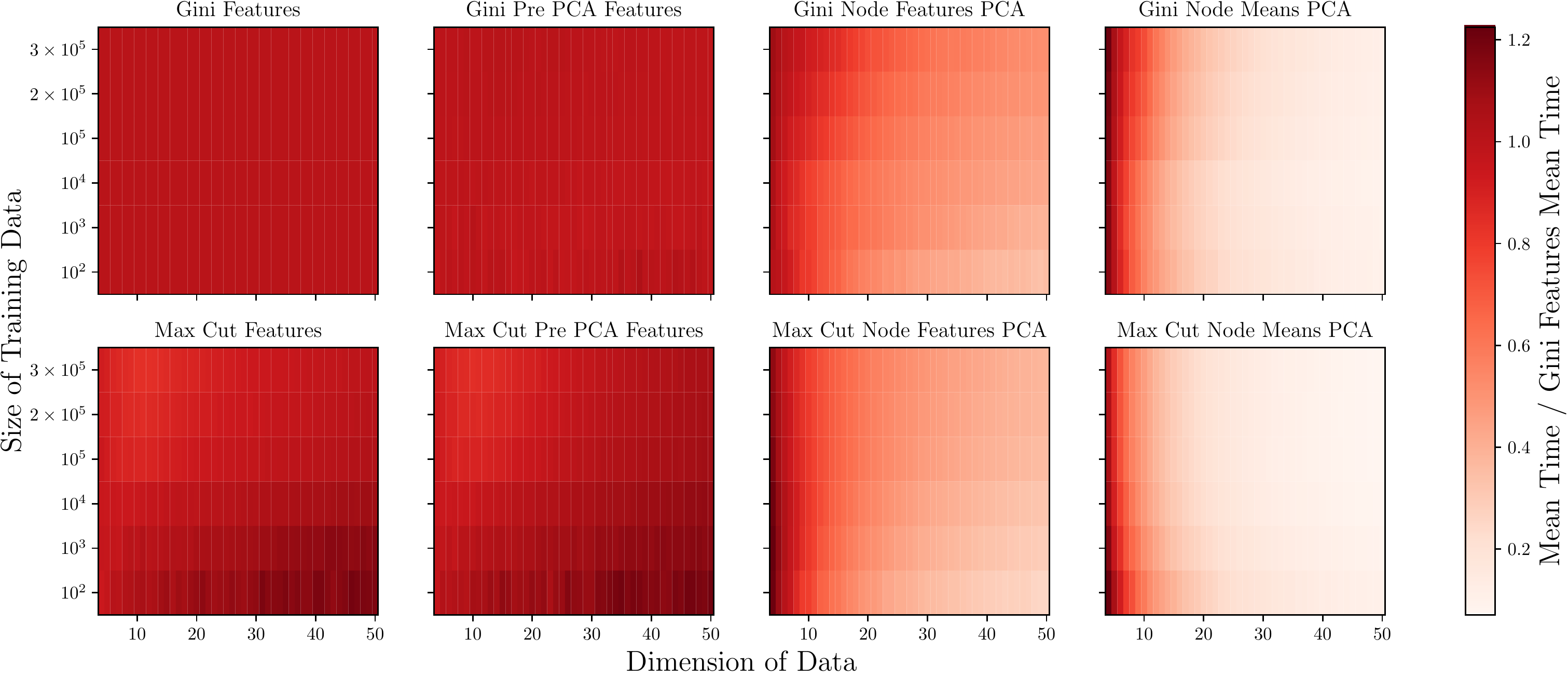, width = 450.pt}}
  \caption{Ratio of mean wall clock time compared to Gini Features baseline for decenary classification decision tree construction on synthetic data.}
  \label{fig:Time of Decenary Classification on Synthetic Data}
\end{figure*}

When considering the results presented in Figure 5, it is noteworthy that much like in the binary case (section 3.2), the prevalence of `\textbackslash\textbackslash' markings in the Gini Features, Gini Pre PCA Features, Max Cut Features, Max Cut Pre PCA Features, and Max Cut Node Features PCA indicates that with statistical significance these are not going to be the best algorithm to choose on average. Moreover, the `//' markings indicate that in most cases, the Max Cut Nodes Means PCA was the best option to choose from an accuracy standpoint. Furthermore, Figure 6 shows that once again the Max Cut Nodes Means PCA provides significant computational advantages, even while providing this increase in accuracy.  

Upon close inspection of Figure 5, the markings indicate that the Gini Node Features algorithm may be the better option accuracy wise for the larger training sets and low dimensions. This is the same pattern that was seen to emerge in the binary case as well; therefore, Figure 7 was generated to show a closer look at the comparison of these two algorithms, specifically zooming in on the 300,000 case and demonstrates that while Gini Node Features may out-perform in the low dimensional, large training set case, Max Cut Node Means PCA rapidly begins to dramatically out-perform Gini Node Features as the number of dimensions increases. 

\begin{figure}[!h]
  \centering
   {\epsfig{file = 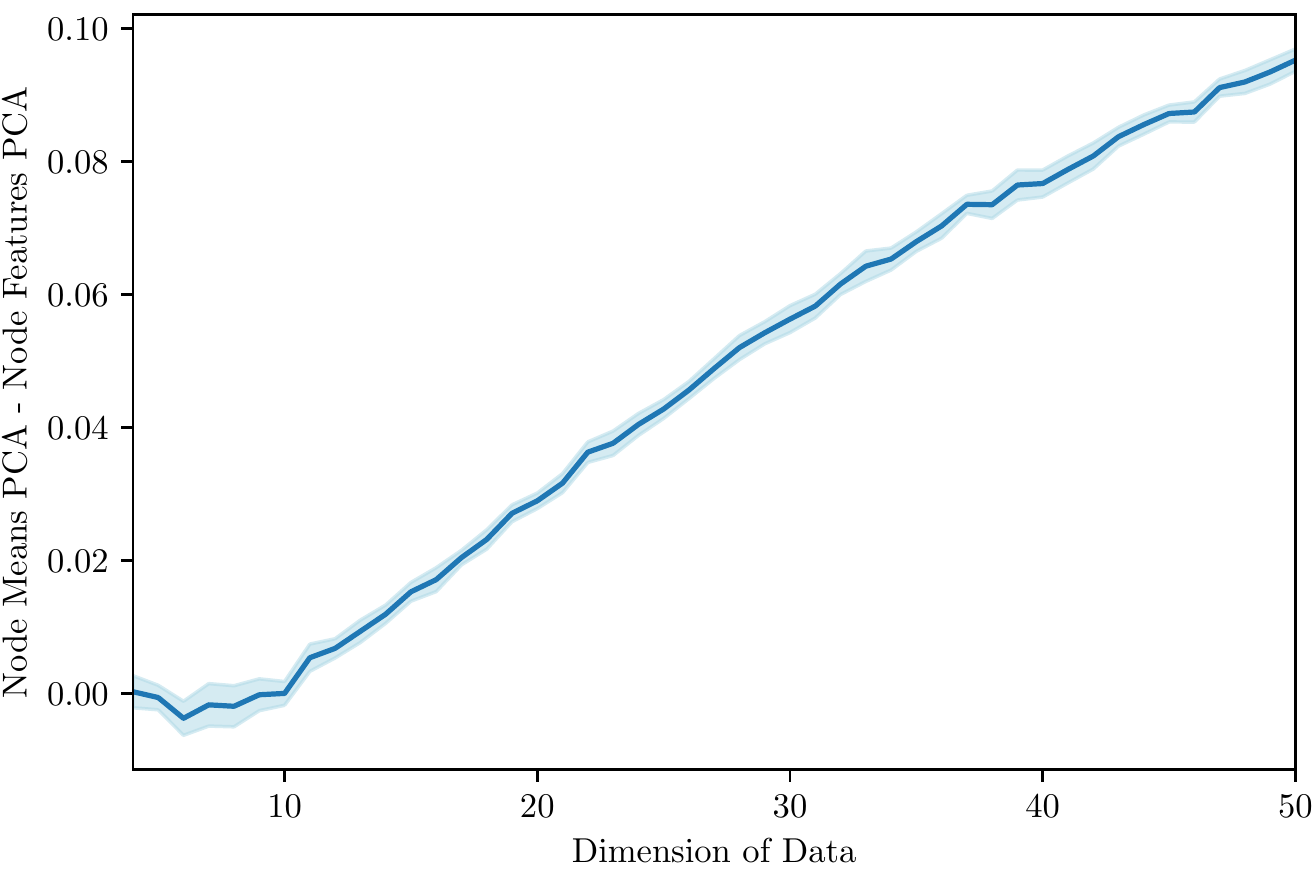, width = 200.pt}}
  \caption{Mean Max Cut Node Means PCA Accuracy minus Mean Gini Node Features PCA Accuracy with 95\% confidence interval using the 300,000 Training Sets.}
  \label{fig:7}
\end{figure}

When the comparison is made between the Max Cut Node Means PCA algorithm and the Gini Node Means PCA algorithm, Figure 8 helps draw a conclusion on the marginal, but still significant improvement by utilizing the Max Cut metric. Specifically, the 95\% confidence interval for the mean ratio between the accuracy of the Max Cut Node Means PCA algorithm and the Gini Node Means PCA algorithm is $(1.0060, 1.0071)$. Therefore, we can conclude that the use of the Max Cut does appear to improve accuracy on average. 

\begin{figure}[!h]
  \centering
   {\epsfig{file = 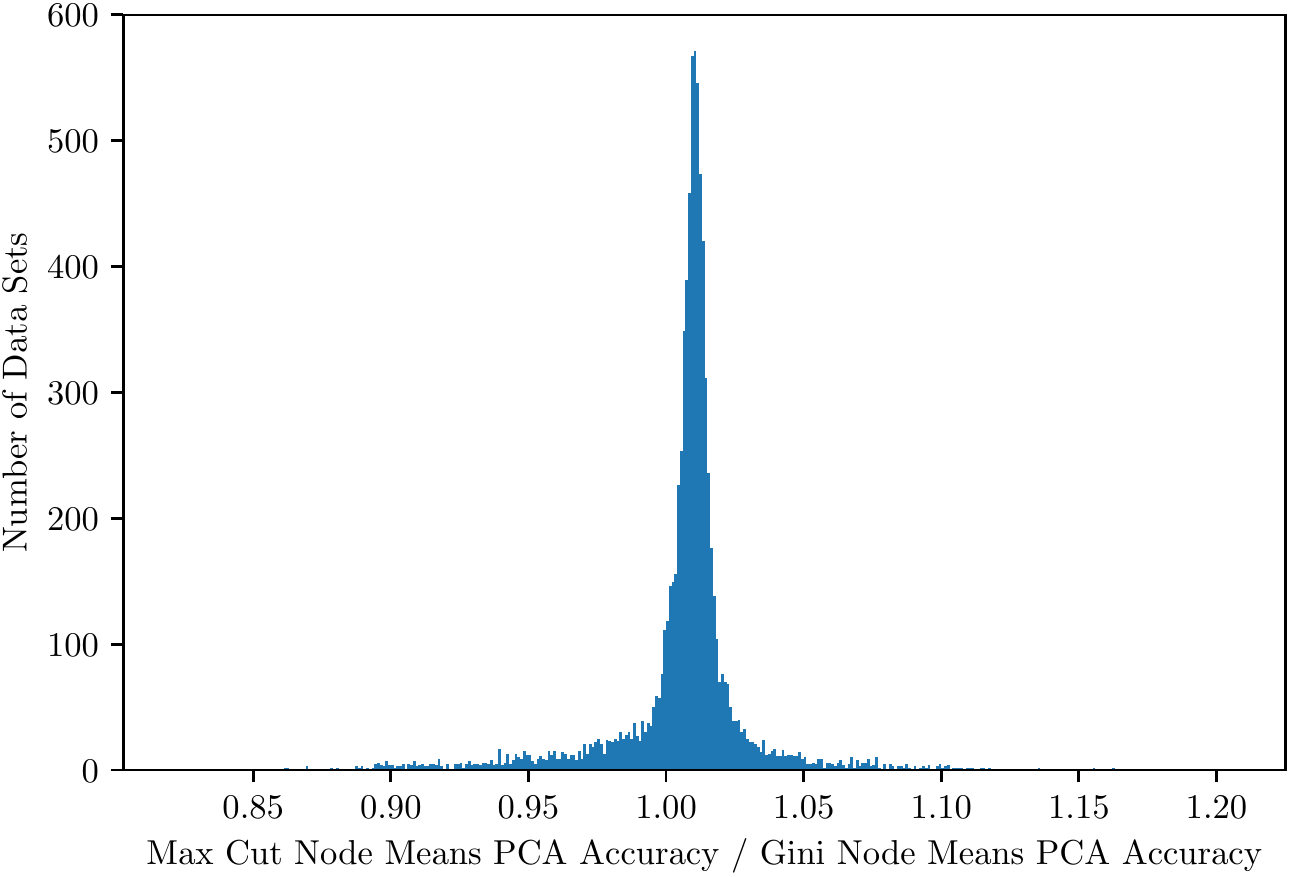, width = 200.pt}}
  \caption{Histogram of Max Cut Node Means PCA Accuracy / Gini Node Means PCA Accuracy.}
  \label{fig:8}
\end{figure}

When analyzing how the performance of the algorithms were affected by the inclusion of noise features, the same results appear, like in the binary analysis, that for low dimensional data (informative dimensions of 5 and 10) and high amount of noise (noise dimensions greater than or equal to 5 and 25 respectfully), the baseline model becomes the better one to chose and our model under-performs. However, these are cases with a very high percentages of noise features (greater than 50\%) and low dimensions. Since the improved accuracy of Max Cut Node Means PCA still held for higher dimensions even when the percentage of noise features exceeded 50\%, we believe that our Max Cut Node Means PCA algorithm is broadly applicable. 

\subsubsection{Synthetic Data Set Conclusions}

After extensive testing, it becomes apparent that our localized PCA algorithms provide dramatic improvements over the baseline model in all but the most extreme noise conditions. A thorough analysis as to which of these localized PCA algorithms should be chosen is provided, finding that from an accuracy standpoint, Node Means PCA should be preferred as the difficulty of the classification task increases, especially when there are more classes, higher dimensions, or fewer training examples.  Moreover, when Node Means PCA was not the best option, it was usually only by a small margin, but can end up dramatically improving results when it is the better option. Empirically, Node Means PCA results in significantly faster computation times compared to the baseline method and the Node Features method. Therefore, the utilization of Node Means PCA provides both dramatic accuracy and run time improvements.

We also analyzed the utilization of the Max Cut metric as an alternative to the Gini Impurity metric and found that when the Max Cut metric was used in conjunction with the Node Means PCA algorithm, it can increase the decision tree's performance on average. Based on our experimentation on more than 20,000 data sets of varying characteristics, we recommend that, in general, Max Cut Node Means PCA be considered as an alternative to the traditional baseline model. While we have shown that Max Cut Node Means PCA yields a significant and dramatic improvement over the baseline model in accuracy and computational efficiency, we understand that our algorithm should be studied under real-world data sets, which is done next.

\subsection{Real-World Data Sets}

We then proceeded to examine the results of the eight different algorithms on real-world data sets. We considered a total of 5 different data sets. We used the Iris Data Set as well as the Wine Quality Data Set \cite{Cortez2009ModelingWP}, both collected from the UCI Machine Learning Repository \cite{Dua:2019}. The Wine Quality Data Set is separated into red and white wines, in our experimentation we consider each of these sets separately and as a single combined data set where the wine's classification as red or white was included as another feature, one if red, zero otherwise. The next data set considered was the MNIST data set \cite{lecun2010mnist}. Finally, both the CIFAR-10 and the CIFAR-100 data sets \cite{Krizhevsky09learningmultiple} were used. Table 4 provides a summary of the different characteristics of each of these data sets.

\begin{table}[h]
\caption{Data Set Characteristics.}\label{tab:4} \centering
\begin{tabular}{|c|ccc|}
  \hline
  Data Set & Samples & Features & Classes \\
  \hline
  Iris & 150 & 4 & 3 \\
  Wine-Red & 1,599 & 11 & 6 \\
  Wine-White & 4,898 & 11 & 7 \\
  Wine-Both & 6,497 & 12 & 7 \\
  MNIST & 70,000 & 784 & 10 \\
  CIFAR-10 & 60,000 & 3,072 & 10 \\
  CIFAR-100 & 60,000 & 3,072 & 100 \\
  
  \hline
\end{tabular}
\end{table}

All of these experiments on real-world data were run on server nodes that each contained two Intel Xeon E5-2670 v2 CPUs for a total of 20 cores (no hyper-threading was used). Each decision tree was constructed using a full node utilizing parallel processing due to the NumPy vector operations. Instead of the wall clock time being record, as in the previous synthetic data set experiments, we recorded the total CPU time for these experiments. 

For all of the algorithms, we considered both when the features are presented to the algorithm without modification as well as when they were standardized. We then reported for each algorithm the best result. To evaluate the performance of each algorithm we used one of two methods: either $10 \times 10$ cross-validation (as in the case of Iris, Wine-Red, Wine-White, and Wine-Both), or 80\% of the data for training and 20\% for testing (as in the case of MNIST, CIFAR-10, and CIFAR-100). The decision to not use $10 \times 10$ cross-validation for MNIST, CIFAR-10, and CIFAR-100 was made due to the significant computational requirements for computing these trees and the fact that due to the large size of the data sets the variance in the accuracy should be lower than in the smaller data sets, this means that the $10 \times 10$ cross-validation would provide less of a benefit. The results of these experiments are reported in Table 5, with the mean value provided for the Iris, Wine-Red, Wine-White, and Wine-Both data sets.

\begin{table*}[!h]
\caption{Results of Real-World Data Set Experiments.}\label{tab:5} \centering
\begin{tabular}{|c|ccc|ccc|}
  \hline
   & \multicolumn{3}{c|}{Iris} &  \multicolumn{3}{c|}{Wine-Red} \\
    & Accuracy & Time & Scaled & Accuracy & Time & Scaled \\
  \hline
  Gini Features & 0.945 & 0.009 & False & 0.626 & 01.5 & False \\
  Gini Pre PCA Features & 0.941 & 0.009 & True & 0.636 & 04.8 & True \\
  Gini Node Features PCA & 0.944 & 0.012 & True & 0.619 & 17.5 & True \\
  Gini Node Means PCA & 0.950 & 0.008 & False & {\bf 0.638} & 10.1 & True \\
  Max Cut Features & 0.947 & 0.010 & True & 0.629 & {\bf 01.2} & True \\
  Max Cut Pre PCA Features & 0.950 & 0.009 & False & 0.632 & 05.0 & True \\
  Max Cut Node Features PCA & 0.941 & 0.011 & False & 0.636 & 18.7 & True \\
  Max Cut Node Means PCA & {\bf 0.960} & {\bf 0.007} & False & 0.631 & 08.3 & True \\
  \hline
   & \multicolumn{3}{c|}{Wine-White} &  \multicolumn{3}{c|}{Wine-Both} \\
    & Accuracy & Time & Scaled & Accuracy & Time & Scaled \\
  \hline
  Gini Features & 0.623 & 06.7 & False & 0.624 & 0:08.5 & False \\
  Gini Pre PCA Features & 0.615 & 08.2 & True & 0.617 & 0:10.4 & False \\
  Gini Node Features PCA & 0.615 & 55.6 & True & 0.608 & 1:17.9 & True \\
  Gini Node Means PCA & {\bf 0.635} & 25.4 & True & 0.633 & 0:36.1 & True \\
  Max Cut Features & 0.627 & {\bf 03.7} & True & 0.623 & {\bf 0:05.5} & True \\
  Max Cut Pre PCA Features & 0.628 & 07.2 & True & 0.626 & 0:08.9 & True \\
  Max Cut Node Features PCA & 0.634 & 58.6 & True & 0.633 & 1:23.3 & True \\
  Max Cut Node Means PCA & {\bf 0.635} & 29.8 & True & {\bf 0.636} & 0:36.3 & True \\
  \hline
    & \multicolumn{3}{c|}{MNIST} &  \multicolumn{3}{c|}{CIFAR-10} \\
    & Accuracy & Time & Scaled & Accuracy & Time & Scaled \\
  \hline
  Gini Features & 0.869 & 0:53:46 & False & 0.262 & 4:21:06 & False \\
  Gini Pre PCA Features & 0.823 & 1:23:46 & True & 0.246 & 5:03:29 & False \\
  Gini Node Features PCA & 0.863 & 2:20:26 & False & 0.290 & 5:06:10 & False \\
  Gini Node Means PCA & 0.922 & 0:10:01 & False & 0.342 & 0:22:10 & False \\
  Max Cut Features & 0.847 & 0:40:01 & False & 0.243 & 5:42:54 & False \\
  Max Cut Pre PCA Features & 0.866 & 0:57:24 & False & 0.271 & 6:32:24 & False \\
  Max Cut Node Features PCA & 0.897 & 1:23:28 & False & 0.309 & 4:30:06 & True \\
  Max Cut Node Means PCA & {\bf 0.924} & {\bf 0:05:03} & False & {\bf 0.349} & {\bf 0:19:35} & False \\
  \hline
      & \multicolumn{3}{c|}{CIFAR-100} &  \multicolumn{3}{c|}{} \\
    & Accuracy & Time & Scaled &  &  &  \\
  \hline
  Gini Features & 0.083 & 24:26:46 & False & & & \\
  Gini Pre PCA Features & 0.073 & 30:46:26 & False & & & \\
  Gini Node Features PCA & 0.090 & 24:41:31 & True & & & \\
  Gini Node Means PCA & 0.116 & 03:13:13 & False & & & \\
  Max Cut Features & 0.077 & 15:18:12 & False & & & \\
  Max Cut Pre PCA Features & 0.090 & 15:34:27 & False & & & \\
  Max Cut Node Features PCA & 0.109 & 12:00:03 & False & & & \\
  Max Cut Node Means PCA & {\bf 0.124} & {\bf 01:29:50} & True & & & \\
  \hline
  
\end{tabular}
\end{table*}

The results of the real-world analysis shows the improvements in accuracy are similar to those from the synthetic data sets. In the case of the Node Means PCA modification, we found that this was the best option to use in all 7 out of 7 problems we analyzed representing an increase in performance (measured as the percentage improvement of accuracy compared to the baseline model after selecting the better of Max Cut and Gini) of 1.6\%, 1.9\%, 1.9\%, 1.9\%, 6.3\% 33.3\%, and 49.4\% for each of the data sets. Moreover, Max Cut was the best choice in 5 out of 7 problems representing an increase in performance (measured as the percentage improvement of the best Max Cut algorithm compared to the best performing Gini algorithm) of 1.1\% -0.3\%, 0.0\%, 0.5\%, 0.2\% 2.0\%, and 6.9\%. Finally, Max Cut Node Means PCA was the best choice in 5 out of 7 problems and further tied for best in 1 out of 7 problems representing an increase in performance (measured as the percentage improvement of accuracy compared to the baseline model) of 1.6\%, 1.0\%, 1.9\%, 1.9\%, 6.3\%, 33.3\%, and 49.4\%.

From Table 5, it can be see that Max Cut Node Means PCA is always the fastest algorithm whenever we consider problems that took the baseline method more that 1 minute of CPU time to compute. Although our algorithm, Max Cut Node Means PCA, took longer in these small data set cases, we do not believe this is of importance since it was never more than 31 CPU seconds slower than the fastest method (while providing the accuracy benefits noted above). For the larger data sets, such as CIFAR-100, the Max Cut Node Means PCA algorithm reduced the required CPU time by 94\% compared to the baseline algorithm accounting for a reduction of 22 hours, 56 minutes, and 56 seconds of necessary CPU time, thereby, providing significant computational and accuracy advantages. 

\section{\uppercase{Conclusions}}

In this paper, we propose two important modifications to the CART methodology for constructing classification decision trees. We first introduced the utilization of a Max Cut based metric (see appendix for $O(n \log n)$ implementation along a single feature) for determining the optimal binary split as an alternative to the Gini Impurity method. We then introduced the use of localized PCA to determine locally good directions for considering splits. We further introduce a modification to the traditional PCA algorithm, Means PCA, to find good directions for discriminating between classes. We follow this up with a theoretical comment on how these modifications affect the asymptotic bounds of node splitting, showing that Means PCA improves upon the traditional method. 

Our extensive experimental analysis included more than 20,000 syntactically generated data sets with training sizes ranging between 100 and 300,000, and informative dimensions ranging between 4 and 50, considering both binary and decenary classification tasks. These experimentations demonstrate the significant improvements provided by localized Means PCA and Max Cut in increased accuracy and decreased computational time. Furthermore, we show that the accuracy improvements only get more dramatic as either the dimension of the data or the number of classes is increased; and that the run time improvements also improve with the dimension and size of the data sets being considered.  

This initial experimentation on synthetic data was followed by an analysis of real-world data sets. Our results indicate that the Max Cut Node Means PCA algorithm continues to have advantages even after the transfer away from synthetic data sets.  For example, we show that in the case of CIFAR-100 (100 classes, 3,072 dimensions, and 48,000 utilized training observations out of 60,000 total) our algorithm Max Cut Node Means PCA results in a 49.4\% increase in accuracy performance compared to the baseline CART model while simultaneously reducing the CPU time required by 94\%. This novel algorithm helps bring decision trees into the world of big, high dimensional data. Throughout, our experiments show the significant improvements that our Max Cut Node Means PCA algorithm for constructing classification decision trees provides for these types of data sets. Further research on how these novel decision trees affect the performance of ensemble methods may lead to even greater advancements in the area. 

\section*{\uppercase{Acknowledgements}}

\noindent This research used the Savio computational cluster resource provided by the Berkeley Research Computing program at the University of California, Berkeley (supported by the UC Berkeley Chancellor, Vice Chancellor for Research, and Chief Information Officer). This research was supported in part by NSF award No. CMMI-1760102.

\bibliographystyle{apalike}
{\small
\bibliography{main}}

\begin{thebibliography}{}

\bibitem[Breiman et~al., 1984]{breiman1984classification}
Breiman, L., Friedman, J., Stone, C., and Olshen, R. (1984).
\newblock {\em Classification and Regression Trees}.
\newblock The Wadsworth and Brooks-Cole statistics-probability series. Taylor
  \& Francis.

\bibitem[Cortez et~al., 2009]{Cortez2009ModelingWP}
Cortez, P., Cerdeira, A., Almeida, F., Matos, T., and Reis, J. (2009).
\newblock Modeling wine preferences by data mining from physicochemical
  properties.
\newblock {\em Decis. Support Syst.}, 47:547--553.

\bibitem[Dua and Graff, 2017]{Dua:2019}
Dua, D. and Graff, C. (2017).
\newblock {UCI} machine learning repository.

\bibitem[F.R.S., 1901]{doi:10.1080/14786440109462720}
F.R.S., K.~P. (1901).
\newblock Liii. on lines and planes of closest fit to systems of points in
  space.
\newblock {\em The London, Edinburgh, and Dublin Philosophical Magazine and
  Journal of Science}, 2(11):559--572.

\bibitem[Holm, 1979]{10.2307/4615733}
Holm, S. (1979).
\newblock A simple sequentially rejective multiple test procedure.
\newblock {\em Scandinavian Journal of Statistics}, 6(2):65--70.

\bibitem[Krizhevsky, 2009]{Krizhevsky09learningmultiple}
Krizhevsky, A. (2009).
\newblock Learning multiple layers of features from tiny images.
\newblock Technical report.

\bibitem[LeCun et~al., 2010]{lecun2010mnist}
LeCun, Y., Cortes, C., and Burges, C. (2010).
\newblock Mnist handwritten digit database.
\newblock {\em ATT Labs [Online]. Available: http://yann.lecun.com/exdb/mnist},
  2.

\bibitem[Oliphant, 2006]{oliphant2006guide}
Oliphant, T.~E. (2006).
\newblock {\em A guide to NumPy}, volume~1.
\newblock Trelgol Publishing USA.

\bibitem[Pedregosa et~al., 2011]{scikit-learn}
Pedregosa, F., Varoquaux, G., Gramfort, A., Michel, V., Thirion, B., Grisel,
  O., Blondel, M., Prettenhofer, P., Weiss, R., Dubourg, V., Vanderplas, J.,
  Passos, A., Cournapeau, D., Brucher, M., Perrot, M., and Duchesnay, E.
  (2011).
\newblock Scikit-learn: Machine learning in {P}ython.
\newblock {\em Journal of Machine Learning Research}, 12:2825--2830.

\bibitem[Van~Rossum and Drake, 2009]{10.5555/1593511}
Van~Rossum, G. and Drake, F.~L. (2009).
\newblock {\em Python 3 Reference Manual}.
\newblock CreateSpace, Scotts Valley, CA.

\end{thebibliography}

\section*{\uppercase{Appendix}}

\noindent Implementation of Max Cut Metric in $O(n \log n)$:

The first step is to sort the observations in order of ascending order such that $x_{i,j} < x_{k,j}\ \forall i < k$ which it is know can be implemented in $O(n \log n)$ time. It can be seen that there are a total of $n$ possible splits to consider, let the value achieved by the split between $x_{i,j}$ and $x_{k,j}$ be referred to as $\theta_i$. We will now show that given $\theta_{i-1}$, $\theta_{i}$ can be computed in constant time. It is easy to see that the following equality holds:

\begin{align*}
    \theta_{i} & = \theta_{i-1} - \sum_{t<i} \mathds{1}_{y_{i} \neq y_{t}}\left|x_{i,j} - x_{t,j}\right| \\
    & \quad \quad \quad + \sum_{t>i} \mathds{1}_{y_{i} \neq y_{t}}\left|x_{i,j} - x_{t,j}\right| \\
    & = \theta_{i-1} - \sum_{t<i} \mathds{1}_{y_{i} \neq y_{t}}\left(x_{i,j} - x_{t,j}\right) \\
    & \quad \quad \quad + \sum_{t>i} \mathds{1}_{y_{i} \neq y_{t}}\left(x_{t,j} - x_{i,j}\right) \\
    & = \theta_{i-1} + \sum_{t<i} \mathds{1}_{y_{i} \neq y_{t}}x_{t,j} - \mathds{1}_{y_{i} \neq y_{t}}x_{i,j} \\
    & \quad \quad \quad + \sum_{t>i} \mathds{1}_{y_{i} \neq y_{t}}x_{t,j} - \mathds{1}_{y_{i} \neq y_{t}}x_{i,j} \\
    \intertext{Since $\mathds{1}_{y_{i} \neq y_{i}} = 0\ \forall i$:}
    & = \theta_{i-1} + \sum_{t} \mathds{1}_{y_{i} \neq y_{t}}x_{t,j} - \mathds{1}_{y_{i} \neq y_{t}}x_{i,j} \\
    & = \theta_{i-1} + \left( \sum_{t} \mathds{1}_{y_{i} \neq y_{t}}x_{t,j} \right) - x_{i,j}\left(\sum_{t} \mathds{1}_{y_{i} \neq y_{t}} \right)
    \intertext{Since $\sum_{t} \mathds{1}_{y_{i} \neq y_{t}}x_{t,j}$ and $\sum_{t} \mathds{1}_{y_{i} \neq y_{t}}$ are only dependent on the class of observation $i$ these can be calculated once for each class, $c$, and recorded as $S_c$ and $N_c$ respectfully, then:}
    \theta_{i} & = \theta_{i-1} + S_{y_{i}} - x_{i,j}N_{y_{i}}
\end{align*}

\noindent Therefore, the complexity of the split per feature is $O(n \log n)$.

\end{document}